\documentclass[sigconf]{acmart}

\AtBeginDocument{%
  }

\providecommand{\adapter}{\textbf{PRIMA}}

\copyrightyear{2026}
\acmYear{2026}
\setcopyright{cc}
\setcctype{by}
\acmConference[MM '26]{Proceedings of the 34th ACM International Conference on Multimedia}{November 10--14, 2026}{Rio de Janeiro, Brazil}
\acmBooktitle{Proceedings of the 34th ACM International Conference on Multimedia (MM '26), November 10--14, 2026, Rio de Janeiro, Brazil}
\acmDOI{10.1145/3767308.3835063}
\acmISBN{979-8-4007-2213-4/2026/11}

\usepackage{cleveref}
\usepackage{multirow}
\usepackage{subcaption}
\usepackage{float}
\usepackage{balance}

\begin{document}

\title{PRIMA: Pose Refined Inductive Motion Adapter for Zero Shot Social Bodily Behavior Recognition}

\author{Mingcong Ding}
\orcid{0009-0000-3024-2447}
\authornote{Corresponding author.}
\affiliation{%
  \institution{Technical University of Munich}
  \city{Munich}
  \country{Germany}
}
\affiliation{%
  \institution{Munich Center for Machine Learning (MCML)}
  \city{Munich}
  \country{Germany}
}

\email{mingcong.ding@tum.de}

\author{Babette Bühler}
\orcid{0000-0003-1679-4979}
\affiliation{%
  \institution{Technical University of Munich}
  \city{Munich}
  \country{Germany}
}
\affiliation{%
  \institution{Munich Center for Machine Learning (MCML)}
  \city{Munich}
  \country{Germany}
}

\email{babette.buehler@tum.de}

\author{Tim Fütterer}
\orcid{0000-0001-5399-9557}
\affiliation{
  \institution{University of Tübingen}
  \city{Tübingen}
  \country{Germany}
}
\email{tim.fuetterer@uni-tuebingen.de}

\author{Ulrich Trautwein}
\orcid{0000-0003-0647-0057}
\affiliation{
  \institution{University of Tübingen}
  \city{Tübingen}
  \country{Germany}
}
\email{ulrich.trautwein@uni-tuebingen.de}

\author{Enkelejda Kasneci}

\orcid{0000-0003-3146-4484}
\affiliation{%
  \institution{Technical University of Munich}
  \city{Munich}
  \country{Germany}
}
\affiliation{%
  \institution{Munich Center for Machine Learning (MCML)}
  \city{Munich}
  \country{Germany}
}
\email{enkelejda.kasneci@tum.de}

\renewcommand{\shortauthors}{Mingcong Ding, Babette Bühler, Tim Fütterer, Ulrich Trautwein, \& Enkelejda Kasneci}


\begin{abstract}
Recognizing fine-grained bodily behaviors, such as self-grooming, postural shifts, and backchanneling, is a useful intermediate step for computational social signal analysis. But the behaviors that matter vary across social contexts, and retraining a model for each new taxonomy is expensive. Vision-language models (VLMs) can adapt to new categories through text prompts, yet they lack sensitivity to body structure and often confuse subtle postural distinctions. We propose \adapter\ (Pose Refined Inductive Motion Adapter), a lightweight module that injects a body-structure prior into frozen VLMs using human pose as a training-time supervisory signal, requiring no pose input at inference. \adapter\ separates a model-specific projection head from a shared core that encodes transferable body-configuration knowledge; adapting to a new VLM requires retraining only the head. We train \adapter\ on a micro-action dataset and evaluate entirely through zero-shot prompting. The adapter improves recognition on its training domain, transfers to additional VLM backbones at different scales with the core frozen, and generalizes to two unseen target datasets from social psychology and daily activity recognition. In a supporting downstream study, the improved behavior predictions also yield higher correlations with human ratings of rapport and dominance in a collaboration context. Within the scope of the tested models and indoor datasets, the results indicate that pose-guided training can produce body-structure priors that generalize across unseen behavior vocabularies and can be partially reused across VLM backbones while preserving RGB-only inference.
\end{abstract}



\begin{teaserfigure}
  \includegraphics[width=\textwidth]{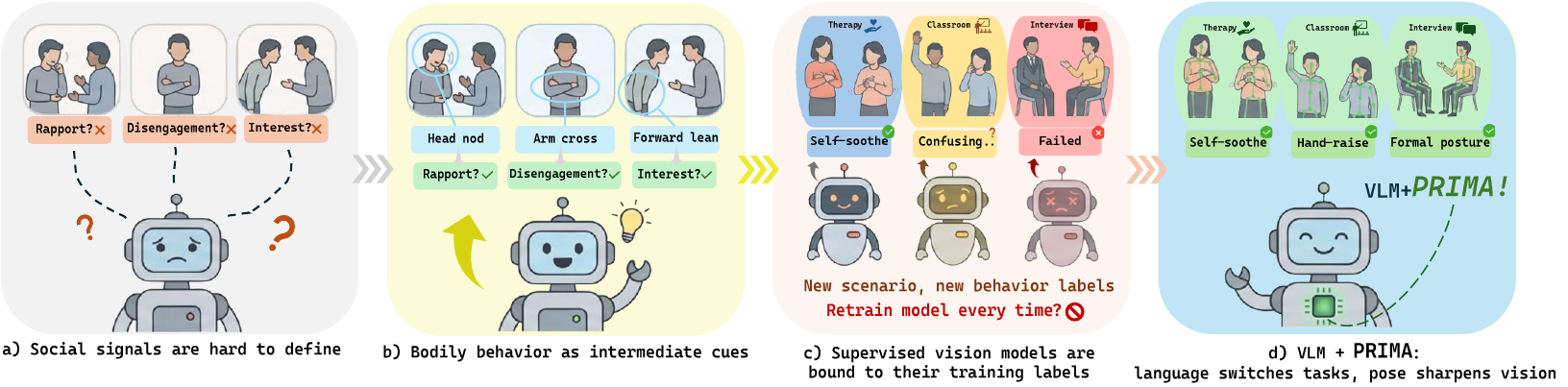}
  \Description{Four-panel illustration. Panel a: two people talking with question-marked labels for rapport, disengagement, and interest. Panel b: the same scene relabelled with concrete bodily behaviors such as head nod, arm cross, and forward lean. Panel c: three separate settings (therapy, classroom, interview) each with its own behavior labels, showing that a supervised model must be retrained for every setting. Panel d: a vision-language model combined with PRIMA recognizing behaviors in all three settings from RGB alone.}
  \caption{From social signals to transferable body structure priors. (a) Social attributes lack stable definitions. (b) Bodily behaviors are concrete, but (c) their label sets change with the setting. (d) \adapter\ injects a pose-guided prior that transfers from RGB alone.}
  \label{fig:teaser}
\end{teaserfigure}

\maketitle

\section{Introduction}
\label{sec:intro}

Understanding subtle social cues embedded in interactions from video matters for healthcare, education, and human robot interaction, but formalizing what to recognize is difficult~\cite{vinciarelli2009social,lee2024towards,vinciarelli2011bridging,rudovic2018personalized}. Most work targets high-level social attributes such as leadership or rapport, yet these lack agreed upon definitions: What counts as rapport varies across disciplines, making annotation unreliable and cross-setting generalization difficult~\cite{sanchez2011nonverbal,tickle1990nature,muller2019emergent,gatica2009automatic}. A different approach focuses on what people \textit{physically do}: head nods, hand gestures, forward leans, postural shifts~\cite{ekman1969repertoire, kendon2004gesture, argyle2013bodily}. Psychology and nonverbal communication research has established that these fine-grained bodily behaviors function as social signals in their own right ~\cite{troisi1999ethological, knapp1972nonverbal, mehu2012psycho, poggi2012social}. Because they are grounded in observable movement, they are easier to define, easier to annotate, and less dependent on contested constructs~\cite{balazia2022bodily, oliveros2015study, troisi1999ethological}. Recent work in social signal processing has organized such behaviors into structured category systems that serve as an intermediate representation between raw video and social interpretation ~\cite{balazia2022bodily, muller2023multimediate, muller2021multimediate, burgoon2017social}.
 
This intermediate representation is useful, but the specific behaviors that matter are not universal. A conversation study may track back channeling and gaze aversion~\cite{amer2023backchannel, muller2023multimediate}; a clinical assessment may focus on self-soothing and repetitive movements~\cite{goodwin2011automated, rajagopalan2013self}. Each context defines its own categories. Existing fine-grained action recognition and bodily behavior recognition methods train supervised classifiers on a fixed label set ~\cite{feichtenhofer2019slowfast, guo2024benchmarking, balazia2022bodily}, so when categories change, the model must be retrained on new data. The cost scales with the number of scenarios, making the approach as expensive as directly predicting high-level attributes. This raises a concrete question: \textit{\textbf{Can we learn general knowledge about how bodies move from one annotated source and transfer it to recognize behaviors under a completely different set of categories, without retraining?}}
 
Vision language models offer a path forward ~\cite{radford2021learning, jia2021scaling, zhang2024vision}. They classify visual inputs solely from natural language descriptions, so when behavior categories change, only the text descriptions need updating ~\cite{radford2021learning}. But their training data overwhelmingly describes objects and scenes at a coarse level, and the subtle postural and gestural differences that distinguish fine-grained behaviors are rarely captured.
Fine-tuning on domain-specific action data risks impairing the general visual knowledge that makes zero-shot transfer possible~\cite{xu2016multi,rohrbach2011evaluating,kumar2022fine}. 

These behaviors are defined by the spatial configuration of body parts, which VLMs do not reliably capture from RGB alone. Human pose encodes exactly this structure, is invariant to clothing and background, and can be extracted automatically from any frame. Existing pose-enhanced VLM methods, however, either require pose at inference or depend on target-task classification labels~\cite{cao2019openpose,zhang2024pevl}. 

We propose \adapter\ (Pose-Refined Inductive Motion Adapter), which injects body structure knowledge into frozen VLMs by aligning their visual features toward pose-informed representations during training. No pose input is needed at inference, and no classification labels are used at any point. The architecture separates a shared core from a VLM-specific projection head; transferring to a new backbone requires retraining only the head while the core stays fixed.

We evaluate entirely through zero-shot VLM prompting. \adapter\ improves recognition on its 52-class training domain~\cite{li2025mmad,guo2024benchmarking}, transfers to a different VLM with the shared core frozen~\cite{bai2025qwen3, wang2025internvl3}, and improves performance on two unseen target datasets: BBSI~\cite{balazia2022bodily} with 14 classes from group interactions and a 30-class subset of NTU~RGB+D~120~\cite{liu2019ntu,shahroudy2016ntu} drawn from daily and health-related activities. In a supporting analysis, the adapter's predictions also yield higher correlations with BBSI's human ratings of rapport and dominance.
 
Our goal is not state-of-the-art absolute scores, but a body-structure prior that transfers across changes in the behavior taxonomy or VLM backbone.

We make three contributions: 
\begin{itemize}
    \item We introduce \adapter, a lightweight adapter that uses human pose as a training-time teaching signal to inject a body-structure prior into frozen VLMs. No pose input is needed at inference, and no classification labels are used during adapter training.
    \item We factor the adapter into a VLM-specific Projection Head and a shared Core, and show that the learned Core can be reused across additional VLM backbones at different scales while retraining only the Head.
    \item We show that a prior learned on MMA~52 improves zero-shot recognition on two unseen target taxonomies, and we analyze where the transfer succeeds (spatial body-part relations) and where it remains limited (fine hand detail and temporal trajectory).
\end{itemize}

\begin{figure*}[t]
\centering
\includegraphics[width=0.75\textwidth]{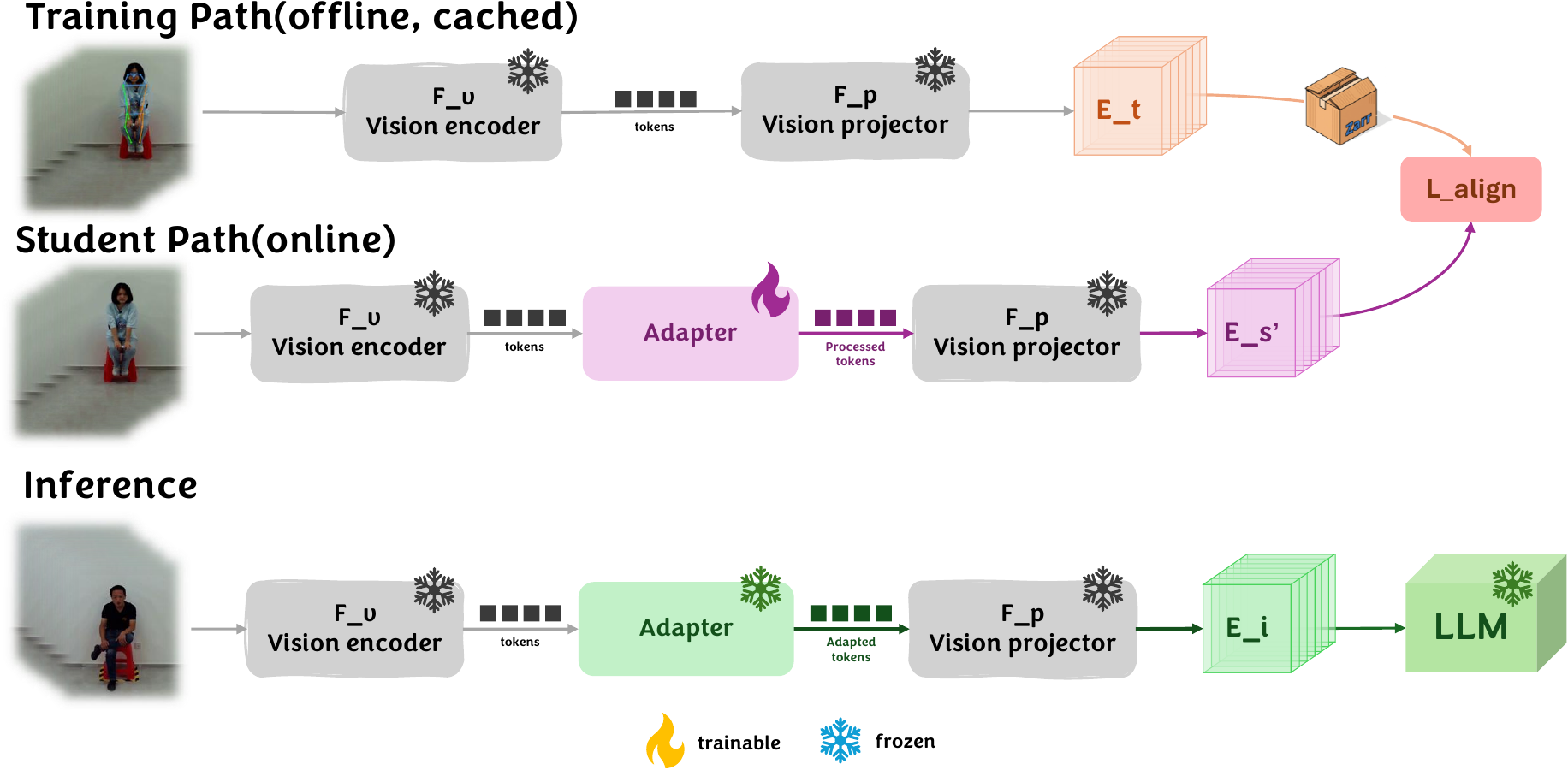}
\Description{Three-row pipeline diagram showing teacher path, student path, and inference path.}
\caption{PRIMA pipeline. \textit{Top}: skeleton overlaid frames produce teacher features $\mathbf{E}_t$ (precomputed, cached). \textit{Middle}: RGB frames pass through the trainable adapter to produce $\mathbf{E}_s'$; $\mathcal{L}_{\text{align}}$ closes the gap. \textit{Bottom}: inference uses only the student path with the adapter frozen.}
\label{fig:archi}
\end{figure*}

\section{Related Work}
\label{sec:related}

\subsection{From Social Attributes to Bodily Behaviors}
Social signal processing has studied how nonverbal behavior encodes social information in group interactions~\cite{vinciarelli2009social, vinciarelli2011bridging, burgoon2017social, beyan2023co}, including gaze-based joint-attention estimation from third-person social-scene video~\cite{sumer2020attention}. Early computational work predicted high-level social attributes from handcrafted features~\cite{gatica2009automatic, hung2010estimating, jayagopi2009modeling}: ~\citet{sanchez2011nonverbal} combined speaking activity and prosody for emergent leader identification, ~\citet{beyan2017moving,beyan2017prediction} applied pose-based features for leadership detection, and ~\citet{muller2018detecting} used facial action units and hand motion for rapport detection. More recent work has estimated student engagement from attentional and affective facial-video features in authentic classroom settings~\cite{sumer2021multimodal,goldberg2021attentive}. M\"uller and Bulling later attempted cross-dataset leadership transfer and observed substantial performance drops~\cite{muller2019emergent}. More recent multimodal work has assessed similarly context-specific constructs such as classroom discourse quality by combining transcript, audio, and video representations~\cite{hou2025multimodal}. Each study defined its own construct, collected setting-specific annotations, and trained a model that does not transfer to other settings. More recently, annotation efforts have shifted toward mid-level bodily behaviors as a more reusable intermediate representation. ~\citet{balazia2022bodily} introduced BBSI, annotating 14 body language classes from the Ethological Coding System for Interviews and evaluating supervised detectors. ~\citet{liu2021imigue} annotated 32 micro gesture classes in iMiGUE, and ~\citet{guo2024benchmarking, li2025mmad} constructed MMA with 52 micro action classes. Each dataset targets subtle body motion, but their label sets come from different traditions and share only a few body configurations while most classes differ. The recognition models are closed-set supervised classifiers that cannot serve as a general foundation across interaction settings~\cite{rajagopalan2013self, noroozi2017audio, muller2024multimediate}. To our knowledge, the transfer of body structure priors across behavior vocabularies has not been systematically evaluated. Cross-modal distillation and pose-guided pretraining operate within fixed label sets and do not test whether learned priors remain useful when the target taxonomy changes substantially.

\subsection{Pose in Action Recognition}
Human pose has been widely used in action recognition as a complementary modality to RGB. Multi-stream architectures such as PoseConv3D and CTR-GCN fuse skeleton sequences with appearance features, requiring pose at both training and inference ~\cite{yan2018spatial, duan2022revisiting, chen2021channel, shi2019two}. Large-scale benchmarks such as NTU~RGB+D~120~\cite{liu2019ntu} provide diverse action categories spanning daily activities and medical conditions, but recognition models are typically evaluated on the same label set used for training. Pose-guided attention mechanisms such as POET and Pose-DRL similarly depend on pose at deployment ~\cite{das2020vpn, das2021vpn++, reilly2023seeing}. Cross-modal distillation removes this dependency: A pose teacher trains an RGB student that operates independently at test time, as in PoseDistill~\cite{hong2021video, thoker2019cross, garcia2018modality} and VPN++~\cite{das2021vpn++}, which distills pose-driven attention into an RGB network for daily activity recognition. But these approaches transfer knowledge within a single dataset and label set, producing no representation intended to generalize across behavior vocabularies~\cite{hinton2015distilling}.

\subsection{Adapting Vision Language Models}
Vision-language models learn joint image-text representations that enable zero-shot classification via natural-language prompting ~\cite{radford2021learning, jia2021scaling, li2023blip}. This open-vocabulary capability is relevant to the cross-taxonomy problem because changing the behavior categories requires only updating the text descriptions, not retraining the model ~\cite{wu2024towards,zhu2024survey}. Substantial work has explored adapting these models to specific visual tasks while preserving generality: CoOp and CoCoOp optimize learnable prompt tokens~\cite{zhou2022learning, zhou2022conditional, khattak2023maple}, LoRA injects trainable low-rank matrices~\cite{hu2022lora}, and adapter-based approaches such as CLIP-Adapter insert small bottleneck layers~\cite{gao2024clip, houlsby2019parameter, chen2022adaptformer}. For video understanding, ST-Adapter and AIM add temporal modeling to image-level backbones~\cite{pan2022st, yang2023aim, ni2022expanding, ju2022prompting}, and ViFi-CLIP~\cite{rasheed2023fine, wang2021actionclip} explores fine-tuning strategies that balance temporal adaptation with retention of pretrained knowledge. These methods share a common pattern: adaptation targets a specific label set, and the resulting model is evaluated on the same or a closely related set~\cite{he2021towards, jia2022visual}, not on unseen behavior vocabularies. PeVL~\cite{zhang2024pevl} uses pose to enhance VLMs for fine-grained action recognition through cross-modality refinement and contrastive learning. It shares our intuition that VLMs lack sensitivity to body structure, but differs in its training objective and transfer scope: PeVL trains on target-task classification labels and requires pose at inference, tying the learned representation to a fixed taxonomy. \adapter\ uses no classification labels, requires no pose at inference, and is evaluated on social behavior categories it has never seen.

\section{Methodology}
\label{sec:method}

The goal of this work is to learn a transferable body structure prior which generalizes across behavior taxonomies without task-specific retraining. The key mechanism is feature alignment guided by pose. When a 2D skeleton is rendered on top of a video frame, a frozen VLM naturally produces features more sensitive to body structure; we treat these as teacher features. A lightweight adapter is trained to map the VLM's features from plain RGB to the pose-informed teacher features. After training, the skeleton branch is discarded, and the adapter operates on RGB alone (\Cref{fig:archi}). \Cref{tab:notation} summarizes the notation used throughout this section.


\begin{table}[t]
\centering
\caption{Notation summary.}
\label{tab:notation}
\footnotesize
\begin{tabular}{@{}cl@{}}
\toprule
Symbol & Description \\
\midrule
$f_v$, $f_p$ & Frozen vision encoder and vision projector \\
$\mathcal{A}$ & PRIMA adapter ($= \mathcal{H} + \mathcal{C}$) \\
$\mathcal{C}$ & Core (model agnostic, operates in $\mathbb{R}^{d_0}$) \\
$\mathcal{H}$ & Projection Head (model specific) \\
$W_{\text{in}}$, $W_{\text{out}}$ & Linear projections in $\mathcal{H}$ \\
$g$ & Learnable gate scalar (init 0.01) \\
$d_v$ & Vision token dimension (VLM dependent) \\
$d_0$ & Bottleneck dimension (768) \\
$\mathbf{I}$, $\mathbf{I}^{+}$ & RGB frame, skeleton overlaid frame \\
$\mathbf{V}_s$, $\mathbf{V}_s'$ & Original and adapted vision tokens \\
$\mathbf{E}_t$, $\mathbf{E}_s'$ & Teacher and student post projector features \\
$\boldsymbol{\delta}$ & Adapter residual before addition to $\mathbf{V}_s$ \\
\bottomrule
\end{tabular}
\end{table}

\begin{figure*}[t]
\centering
\includegraphics[width=1.9\columnwidth]{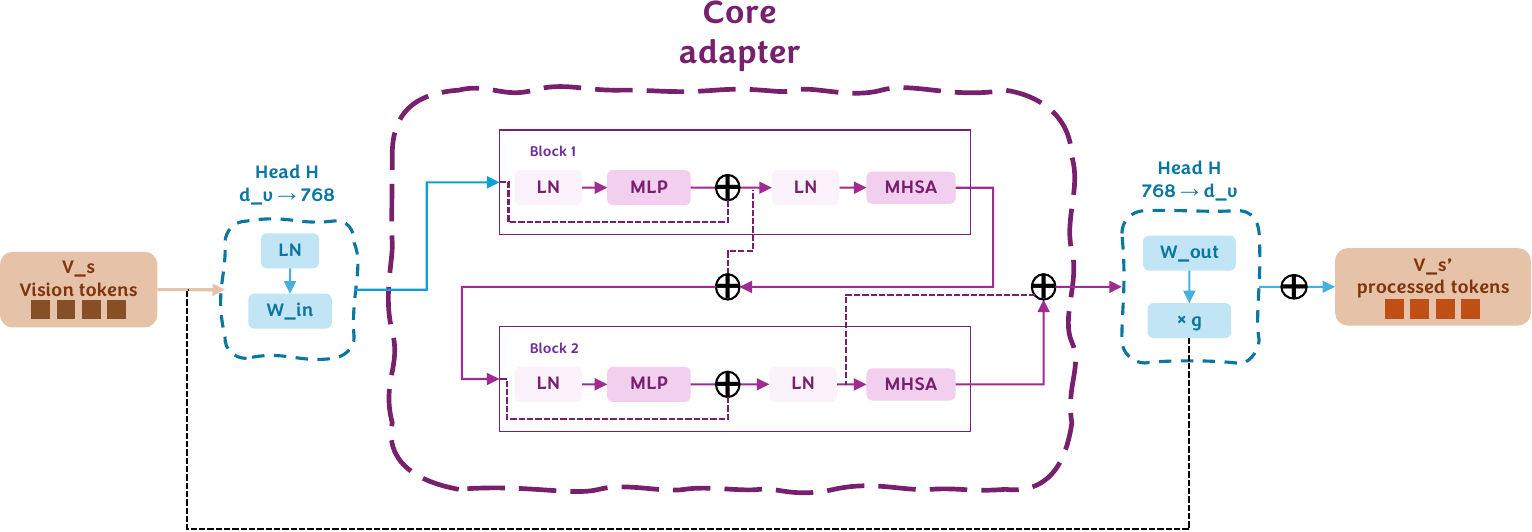}
\Description{Vertical flowchart of the adapter with Head and Core components.}
\caption{Adapter internal structure. Blue: Head $\mathcal{H}$ ($d_v \leftrightarrow 768$, model specific). Purple: Core $\mathcal{C}$ (two blocks, model agnostic). Gate $g$ scales $\boldsymbol{\delta}$ before addition to $\mathbf{V}_s$.}
\label{fig:adapter}
\end{figure*}

\subsection{Pose-guided Feature Alignment}
\label{sec:alignment}

A standard VLM processes a video frame in two stages: a vision encoder $f_v$ converts the image into a sequence of visual tokens, and a vision projector $f_p$ maps those tokens into the language model's embedding space. We add a small trainable adapter $\mathcal{A}$ between the two, keeping the VLM entirely frozen so that the adapter must encode body-structure information as a small, transferable adjustment rather than a large, model-specific distortion.

For each frame $\mathbf{I}$, we use MMPose~\cite{jiang2023rtmpose} to estimate 2D body keypoints and render them as colored dots and bones on the original image, producing a skeleton overlaid frame $\mathbf{I}^{+}$. Feeding $\mathbf{I}^{+}$ through the frozen VLM yields features that retain visual context but are also informed by body structure:
\begin{equation}
\label{eq:teacher}
\mathbf{E}_t = f_p(f_v(\mathbf{I}^{+}))
\end{equation}

Teacher features are computed once and cached to disk (about 150\,GB per backbone). The teacher signal depends on pose estimation quality, but the dependence is mild: with synthetic keypoint noise at 20\% of the person bounding box diagonal, the adapter still reaches 0.278 $\text{F1}_{\text{pool}}$ on MMA~52 against the 0.253 baseline, and performance is insensitive to the rendering style of the overlay. 

On the student side, we feed the same frame without any skeleton overlay through the frozen vision encoder, pass the resulting tokens through the adapter, and then through the frozen projector:
\begin{equation}
\label{eq:student}
\mathbf{E}_s' = f_p(\mathcal{A}(f_v(\mathbf{I})))
\end{equation}

The adapter is trained to make $\mathbf{E}_s'$ match $\mathbf{E}_t$: it learns to reproduce the features the VLM would produce if a skeleton were visible, using only plain RGB. We align in post-projector space because these features directly drive the language model's output.

At inference (\Cref{fig:archi}, bottom row), only the student path is used: each frame passes through frozen encoder, frozen adapter, and frozen projector. No pose input is needed.

\subsection{Adapter Architecture}
\label{sec:architecture}

The adapter sits between the vision encoder output and the vision projector input (\Cref{fig:adapter}). It takes in the original vision tokens $\mathbf{V}_s \in \mathbb{R}^{N \times d_v}$ produced by $f_v$ and outputs adapted tokens $\mathbf{V}_s'$ of the same shape. The dimension $d_v$ depends on the VLM ($d_v = 1152$ for Qwen3 VL, $d_v = 4096$ for InternVL3.5 after pixel shuffle).

\noindent\textbf{Gated residual.}
The adapter computes a residual $\boldsymbol{\delta}$ added to the original tokens:

\begin{equation}
\label{eq:adapter}
\mathbf{V}_s' = \mathbf{V}_s + g \cdot \boldsymbol{\delta}
\end{equation}

where $g$ is a learnable scalar initialized to 0.01, keeping adapted features close to the VLM's original representation at the start of training.

The residual $\boldsymbol{\delta}$ is produced by passing $\mathbf{V}_s$ through two components in sequence, a Projection Head $\mathcal{H}$ and a Core $\mathcal{C}$:
\begin{equation}
\label{eq:delta}
\boldsymbol{\delta} = W_{\text{out}}(\mathcal{C}(W_{\text{in}}(\text{LN}(\mathbf{V}_s))))
\end{equation}

\noindent\textbf{Projection Head $\mathcal{H}$.}
Different VLMs produce vision tokens of different dimensions $d_v$; operating directly in $d_v$ space would tie the adapter to one VLM. The Head maps tokens into a VLM-independent bottleneck: LayerNorm followed by $W_{\text{in}}$ ($d_v \to d_0 = 768$), then after the Core, $W_{\text{out}}$ ($d_0 \to d_v$). Only the Head changes when switching VLMs. The bottleneck at $d_0 = 768$ sits below $d_v$, forcing the Core to learn structured representations.

\noindent\textbf{Core $\mathcal{C}$.}
The Core operates entirely in $\mathbb{R}^{d_0}$ and consists of two identical blocks, each applying an MLP and multi-head self-attention with residual connections:
\begin{align}
\label{eq:core_mlp}
\mathbf{Z} &= \mathbf{X} + \text{MLP}(\text{LN}(\mathbf{X})) \\
\label{eq:core_attn}
\mathbf{X}' &= \mathbf{Z} + \text{MHSA}(\text{LN}(\mathbf{Z}))
\end{align}
The MLP expands to $4d_0$ with GELU activation. The MHSA uses 4 heads, allowing the Core to model spatial dependencies across tokens, which matters because body configuration is relational: Whether a person leans forward depends on the torso's position relative to the legs. Both operations use a dropout rate of 0.1 and the pre-norm convention. Because the Core operates in a fixed dimension independent of $d_v$, it is decoupled from any particular VLM (\Cref{sec:transfer}).

The adapter adds approximately 16M parameters on Qwen3~VL and 20.5M on InternVL3.5 (under 0.3\% of either 8B VLM); the difference comes entirely from the Head.

\subsection{Training Objective}
\label{sec:objective}

The adapter is trained to minimize:

\begin{equation}
\label{eq:total_loss}
\mathcal{L}_{\text{total}} = \mathcal{L}_{\text{align}} + \lambda \, \mathcal{L}_{\text{reg}}
\end{equation}

The alignment loss combines MSE (feature magnitude) and cosine distance (direction):
\begin{equation}
\label{eq:align_loss}
\mathcal{L}_{\text{align}} = \frac{1}{2}\,\text{MSE}(\mathbf{E}_s',\, \mathbf{E}_t) \;+\; \frac{1}{2}\,(1 - \cos(\mathbf{E}_s',\, \mathbf{E}_t))
\end{equation}

The regularization loss prevents the adapter from modifying the original features too aggressively:
\begin{equation}
\label{eq:reg_loss}
\mathcal{L}_{\text{reg}} = \text{mean}(\boldsymbol{\delta}^2), \quad \lambda = 0.01
\end{equation}

This penalizes large residuals, keeping the adapter's adjustments small and the learned prior transferable.

\noindent\textbf{Training details.}
We train for 20 epochs on MMA~52 (4533 clips) using AdamW~\cite{loshchilov2017decoupled} ($\text{lr} = 10^{-4}$, weight decay 0.01), cosine schedule with 10\% warmup, gradient clipping at 1.0, and batch size 1 on a single A100 80GB GPU. Each clip is sampled at 2\,fps (up to 768 frames) at a resolution of $448 \times 448$.

\subsection{Cross-Backbone Transfer}
\label{sec:transfer}

\begin{figure}[t]
\centering
\includegraphics[width=0.7\columnwidth]{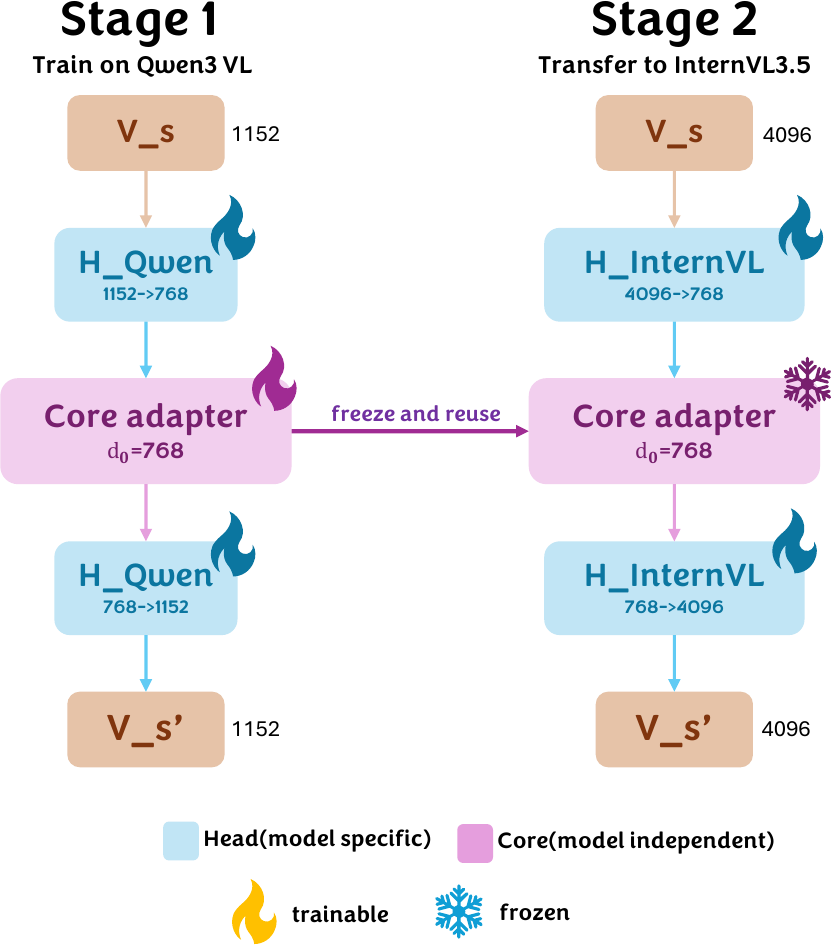}
\Description{Two-stage transfer diagram showing Core reuse across VLMs.}
\caption{Cross backbone transfer. Stage 1 trains the full adapter on Qwen3 VL. Stage 2 freezes the Core and trains only a new Head for InternVL3.5.}
\label{fig:transfer}
\end{figure}

The reason we split the adapter into a Head and a Core is to enable transfer across VLM backbones (\Cref{fig:transfer}).
In Stage 1, we train the full adapter on Qwen3~VL ($W_{\text{in}}$: $1152 \to 768$, $W_{\text{out}}$: $768 \to 1152$). In Stage 2, we freeze the Core and train only a new Head for InternVL3.5, whose token dimension differs ($d_v = 4096$). The new Head ($W_{\text{in}}$: $4096 \to 768$, $W_{\text{out}}$: $768 \to 4096$) maps InternVL tokens into the 768-dimensional space where the Core already operates, then maps the results back.

We validate the design on InternVL3.5 at both 8B and 4B scales, freezing the Core learned on Qwen3~VL and retraining only the Head while keeping the teacher embeddings fixed. Because the new Head must bridge both a dimensional mismatch and a distributional gap between the two backbones, successful transfer indicates that the Core captures body-structure information not tied to a single backbone's feature space. A full adapter trained natively on InternVL3.5 performs slightly better (0.446 versus 0.427 $\text{F1}_{\text{pool}}$ on BBSI), with the frozen Core retaining 68 to 77\% of the native gain across the three datasets. Transfer also holds in the reverse direction: freezing a Core trained natively on InternVL3.5 and training only a new Qwen3~VL Head reaches 0.479 $\text{F1}_{\text{pool}}$ on BBSI against the 0.443 baseline, a gain of 0.036 averaged over three seeds, again retaining about 77\% of the corresponding native gain.


\section{Experiments}
\label{sec:experiments}

\subsection{Experimental Setup}
\label{sec:setup}

All experiments use zero-shot VLM prompting with no target-task labels.

\noindent\textbf{Datasets.}
We use three action recognition datasets drawn from different research areas, whose label sets share only a few body configurations while most classes differ.
MMA~52~\cite{li2025mmad,guo2024benchmarking} contains 52 fine-grained micro action classes covering everyday body movements such as scratching, waving, and nodding. The training split has 4533 video clips, and the test split has 737 samples. Each clip may contain multiple simultaneous actions, making this a multi-label recognition problem.
BBSI~\cite{balazia2022bodily} annotates 14 body language categories from the Ethological Coding System for Interviews on group interaction recordings from the MultiMediate corpus~\cite{muller2021multimediate,muller2023multimediate}, including behaviors such as self-grooming, leg crossing, and postural shifts. The test set contains 947 samples. BBSI is used only for evaluation; neither the adapter nor any VLM sees BBSI data during training.
NTU~RGB+D~120~\cite{liu2019ntu} contains 120 action classes covering daily activities, medical conditions, and mutual interactions. We evaluate on a 30-class single-person subset (20 daily activity classes + 10 medical condition classes), excluding mutual interactions that require two visible people. The test set contains 1000 samples under the cross-subject protocol. NTU~RGB+D~120 is used only for evaluation.

BBSI is the primary transfer target because its categories directly encode subtle actions related to social behavior. NTU serves as a supplementary test whose categories are not social in origin, probing whether the prior generalizes to action recognition beyond the social domain. A few labeled actions partially overlap between MMA~52 and NTU (e.g.\ ``nodding'' and ``nod head/bow''), but the visual domains and annotation granularity differ.

\begin{table}[t]
\centering
\caption{BBSI evaluation under the original frame-level MAP protocol~\cite{balazia2022bodily}. Supervised results are from the original benchmark using Swin features with PDAN detection trained on BBSI labels. Our methods use zero-shot prompting with no BBSI training data.}
\label{tab:bbsi_map}
\footnotesize
\begin{tabular}{@{}lcc@{}}
\toprule
Method & micro MAP & macro MAP \\
\midrule
Swin + PDAN (supervised)~\cite{balazia2022bodily} & 0.745 & 0.374 \\
\midrule
Qwen3 VL baseline (zero-shot) & 0.403 & 0.151 \\
\adapter\ (zero-shot) & \textbf{0.448} & \textbf{0.185} \\
\bottomrule
\end{tabular}
\end{table}

\noindent\textbf{Evaluation protocol.}
We treat each clip as a multi-label classification instance. The model produces a predicted label set $\hat{Y}_i$ for each clip $i$, which is compared against the ground truth label set $Y_i$. We define per-sample counts: $\text{TP}_i = |Y_i \cap \hat{Y}_i|$, $\text{FP}_i = |\hat{Y}_i \setminus Y_i|$, $\text{FN}_i = |Y_i \setminus \hat{Y}_i|$. We report two F1 variants that capture different aspects of performance.

Sample-pooled F1 ($\text{F1}_{\text{pool}}$) aggregates all counts across samples before computing a single score:
\begin{equation}
\label{eq:f1_pool}
\text{F1}_{\text{pool}} = \frac{2 \sum_i \text{TP}_i}{2 \sum_i \text{TP}_i + \sum_i \text{FP}_i + \sum_i \text{FN}_i}
\end{equation}
This reflects overall prediction quality but is influenced more by frequent classes.

Class averaged F1 ($\text{F1}_{\text{cls}}$) computes F1 for each class $c$ independently, then averages over all $C$ classes:
\begin{equation}
\label{eq:f1_cls}
\text{F1}_{\text{cls}} = \frac{1}{C} \sum_{c=1}^{C} \frac{2\,\text{TP}_c}{2\,\text{TP}_c + \text{FP}_c + \text{FN}_c}
\end{equation}
This gives equal weight to every class regardless of frequency.

The three datasets use incompatible native metrics, so we adopt clip-level F1 for cross-taxonomy comparison. We additionally report BBSI and MMA~52 under their native metrics in \Cref{sec:main_results}, confirming that relative gains hold. NTU's native protocol requires training data we do not use, so we rely on clip-level F1.

\noindent\textbf{Zero-shot prompting.}
We split each dataset's label set into three anatomical groups (head, upper body, lower body) and make one VLM call per group, presenting the relevant category names with numeric identifiers; shorter candidate lists reduce the chance that the VLM overlooks or confuses visually similar categories. The same prompt templates, parsing rules, and retry logic are used across all models and all compared methods; only the category names change. We therefore treat prompting as part of the evaluation protocol rather than part of the proposed method. A single prompt instead of grouped prompting lowers absolute scores for all methods by a similar margin without changing their ranking. Retry and union-merging are deterministic post-processing applied identically to all methods; disabling retries changes fewer than 2\% of predictions and does not alter the ranking.

\noindent\textbf{Models.}
We use three VLMs: Qwen3~VL~8B Instruct~\cite{bai2025qwen3}, InternVL3.5~8B Instruct, and InternVL3.5~4B Instruct~\cite{wang2025internvl3}. The adapter is trained on Qwen3~VL using MMA~52 training data. For both InternVL3.5 variants, we freeze the Core learned on Qwen3~VL and train only a new Projection Head (\Cref{sec:transfer}). All three VLMs are frozen throughout; only the adapter parameters are trainable.

\begin{table}[t]
\centering
\caption{Zero-shot recognition results. All methods use the same VLM and prompting protocol with no target-dataset labels. \adapter\ uses pose only during training; at inference, it receives only RGB. Best result per backbone in \textbf{bold}; $\pm$ values across three seeds $\leq 0.005$ on all settings. InternVL3.5 4B tests cross-scale transfer with the same frozen Core.}
\label{tab:main}
\tiny
\setlength{\tabcolsep}{5pt}
\begin{tabular}{@{}ll cc cc cc@{}}
\toprule
& & \multicolumn{2}{c}{MMA~52 (source)} & \multicolumn{2}{c}{BBSI (transfer)} & \multicolumn{2}{c}{NTU (transfer)} \\
\cmidrule(lr){3-4} \cmidrule(lr){5-6} \cmidrule(lr){7-8}
Backbone & Method & $\text{F1}_{\text{pool}}$ & $\text{F1}_{\text{cls}}$ & $\text{F1}_{\text{pool}}$ & $\text{F1}_{\text{cls}}$ & $\text{F1}_{\text{pool}}$ & $\text{F1}_{\text{cls}}$ \\
\midrule
\multirow{2}{*}{Qwen3~VL 8B}
& RGB baseline & 0.253 & 0.162 & 0.443 & 0.219 & 0.391 & 0.243 \\
& \adapter & \textbf{0.324} & \textbf{0.186} & \textbf{0.490} & \textbf{0.232} & \textbf{0.438} & \textbf{0.278} \\
\midrule
\multirow{2}{*}{InternVL3.5 8B}
& RGB baseline & 0.174 & 0.109 & 0.382 & 0.156 & 0.312 & 0.186 \\
& \adapter\ (frozen Core) & \textbf{0.213} & \textbf{0.158} & \textbf{0.427} & \textbf{0.208} & \textbf{0.358} & \textbf{0.227} \\
\midrule
\multirow{2}{*}{InternVL3.5 4B}
& RGB baseline & 0.148 & 0.091 & 0.349 & 0.137 & 0.282 & 0.163 \\
& \adapter\ (frozen Core) & \textbf{0.175} & \textbf{0.121} & \textbf{0.384} & \textbf{0.175} & \textbf{0.316} & \textbf{0.196} \\
\bottomrule
\end{tabular}
\end{table}

\subsection{Main Results}
\label{sec:main_results}

\Cref{tab:main} reports zero-shot recognition results across nine evaluation settings: three VLM backbones and three datasets. Each setting compares the baseline VLM receiving only RGB against \adapter\ receiving only RGB after pose-guided training.

\noindent\textbf{Pose information improves the ability of VLMs to capture fine-grained body motion.}
An Oracle receiving skeleton overlaid frames outperforms the RGB baseline in all six settings we measured on Qwen3~VL and InternVL3.5~8B. On MMA~52 it reaches 0.355 and 0.249 $\text{F1}_{\text{pool}}$ against 0.253 and 0.174 for the RGB baseline; on BBSI it reaches 0.512 and 0.463 against 0.443 and 0.382. These Oracle runs are reported in text only and are not tabulated. The $\text{F1}_{\text{cls}}$ pattern is similar, and the gap confirms that VLMs do not fully capture body structure from RGB alone.

\noindent\textbf{\adapter\ internalizes the pose signal.}
\adapter\ improves over the RGB baseline in all nine settings without receiving any pose input at inference (\Cref{tab:main}). On MMA~52, $\text{F1}_{\text{pool}}$ improves by +28\% on Qwen3~VL (0.253 to 0.324) and +22\% on InternVL3.5~8B (0.174 to 0.213), recovering about 70\% and 52\% of the Oracle gap respectively.

\noindent\textbf{The body structure prior transfers across taxonomies.}

The adapter was trained on MMA~52 only, yet it improves recognition on both unseen target datasets. On BBSI, $\text{F1}_{\text{pool}}$ improves by +11\% on Qwen3~VL (0.443 to 0.490) and +12\% on InternVL3.5~8B (0.382 to 0.427). On NTU, $\text{F1}_{\text{pool}}$ improves by +12\% and +15\% respectively. The $\text{F1}_{\text{cls}}$ gains are proportionally larger, indicating that the adapter helps rare classes as well. A paired bootstrap test (10,000 resamples) confirms significance at $p < 0.01$ on all settings, including InternVL3.5~4B, where the frozen Core still yields relative $\text{F1}_{\text{pool}}$ gains of +10\% to +18\%. A randomly initialized Core paired with a trained Head remains near the baseline, so the trained Core's content, not the Head's alignment procedure, drives transfer.

\noindent\textbf{Which behaviors benefit and which do not.}
Per-class analysis on BBSI shows that behaviors defined by clear spatial joint relationships benefit most (Hand-face +8.2 pp, Legs crossed +6.5 pp on Qwen3~VL), while behaviors requiring fine hand detail (Fumble, Smearing hands) or temporal trajectory (Gesture) degrade slightly, with the same ranking on Qwen3~VL and InternVL3.5~8B. A post-hoc temporal averaging window ($w{=}3$) recovers most of the Gesture drop ($-$2.8 to $-$0.9\,pp) at minimal cost to the spatial categories.

\noindent\textbf{Comparison under the BBSI native evaluation protocol.}
\Cref{tab:bbsi_map} evaluates BBSI under its original frame-level MAP~\cite{balazia2022bodily}; under MMA~52's native temporal detection mAP, \adapter\ improves mAP@0.3 from 0.112 to 0.149. The gap to supervised baselines reflects training on target labels. The relevant finding is that \adapter\ improves over the zero-shot baseline under both native protocols, confirming that the gains are not artifacts of our clip-level evaluation.

\begin{table}[t]
\centering
\caption{Ablation on MMA~52 with Qwen3~VL. Each row modifies one component.}
\label{tab:ablation}
\footnotesize
\begin{tabular}{@{}l cc@{}}
\toprule
Variant & $\text{F1}_{\text{pool}}$ & $\text{F1}_{\text{cls}}$ \\
\midrule
\adapter\ (full) & 0.324 & 0.186 \\
\midrule
\multicolumn{3}{@{}l}{\textit{Loss components}} \\
\quad w/o cosine loss (MSE only) & 0.305 & 0.169 \\
\quad w/o MSE loss (cosine only) & 0.291 & 0.158 \\
\quad w/o regularization ($\lambda = 0$) & 0.313 & 0.175 \\
\midrule
\multicolumn{3}{@{}l}{\textit{Teacher signal}} \\
\quad Skeleton on blank background & 0.278 & 0.147 \\
\quad Scrambled skeleton & 0.244 & 0.113 \\
\midrule
\multicolumn{3}{@{}l}{\textit{Control}} \\
\quad Random adapter (no pose training) & 0.252 & 0.129 \\
\bottomrule
\end{tabular}
\end{table}

\subsection{Ablation Studies}
\label{sec:ablation}
We ablate key design choices on MMA~52 using Qwen3~VL (\Cref{tab:ablation}).

\noindent\textbf{Loss components.}
Removing cosine (MSE only) drops $\text{F1}_{\text{pool}}$ to 0.305: MSE matches feature magnitude but does not constrain direction, so the adapted features can drift in orientation while staying close in norm. Removing MSE (cosine only) drops it further to 0.291, because cosine alignment ignores scale differences that the projector is sensitive to. Removing regularization yields 0.313; without the penalty on $\boldsymbol{\delta}$, the adapter makes larger modifications that fit the source domain more tightly but are harder to transfer. 

\noindent\textbf{Teacher signal.}
Rendering the skeleton on a blank background drops $\text{F1}_{\text{pool}}$ to 0.278: Without visual context, teacher features diverge too far from the VLM's natural distribution. A scrambled skeleton control (keypoints randomly reassigned across body locations) drops to 0.244, confirming that anatomically correct structure is essential.

\noindent\textbf{Random adapter.}
An adapter with identical architecture but random untrained weights scores 0.252, essentially matching the baseline's 0.253. The improvement from \adapter\ comes from pose-guided training, not from the adapter's structure.

\noindent\textbf{Comparison with supervised adaptation.}
\Cref{tab:compare} compares \adapter\ with LoRA~\cite{hu2022lora} (rank 16) and a supervised adapter (same architecture, cross-entropy loss), both trained on MMA~52's 52-class labels. On MMA~52, supervised methods outperform \adapter\ as expected: LoRA leads by +0.038 $\text{F1}_{\text{pool}}$. The comparison reverses on transfer: on BBSI, LoRA drops to 0.451 (near the 0.443 baseline) while \adapter\ reaches 0.490; on NTU, \adapter\ leads by a similar margin. Methods that fit source-label boundaries lose their advantage when the taxonomy changes. We exclude PeVL~\cite{zhang2024pevl} because it requires target-task labels and pose at inference, so a direct comparison would differ in inputs rather than isolate the effect we study. The supervised adapter in \Cref{tab:compare} isolates this factor: same architecture, classification loss, and its BBSI transfer (0.462) barely exceeds the baseline (0.443), confirming that the training signal, not the architecture, limits cross-taxonomy transfer.

Two further controls isolate what drives transfer. An augmentation alignment baseline with the same architecture and loss but a pose-free teacher reaches 0.454 $\text{F1}_{\text{pool}}$ on BBSI against 0.490 for \adapter, so the alignment procedure alone does not explain the gain. A Core with temporal instead of spatial attention, all else fixed, transfers worse (+4.3\% over the baseline on BBSI versus +11\% for \adapter), indicating that spatial body-part modeling rather than temporal dynamics carries the prior across taxonomies.

\begin{table}[t]
\centering
\caption{Comparison with alternative adaptation methods on Qwen3~VL. All methods use the same zero-shot prompting protocol at evaluation. Supervised methods are trained on MMA~52 classification labels. \adapter\ uses no classification labels.}
\label{tab:compare}
\footnotesize
\setlength{\tabcolsep}{4pt}
\begin{tabular}{@{}l l cc@{}}
\toprule
Dataset & Method & $\text{F1}_{\text{pool}}$ & $\text{F1}_{\text{cls}}$ \\
\midrule
\multirow{4}{*}{MMA~52}
& RGB baseline & 0.253 & 0.162 \\
& LoRA (supervised) & \textbf{0.362} & \textbf{0.214} \\
& Supervised adapter & 0.341 & 0.198 \\
& \adapter\ (pose-guided) & 0.324 & 0.186 \\
\midrule
\multirow{4}{*}{BBSI}
& RGB baseline & 0.443 & 0.219 \\
& LoRA (supervised) & 0.451 & 0.215 \\
& Supervised adapter & 0.462 & 0.221 \\
& \adapter\ (pose-guided) & \textbf{0.490} & \textbf{0.232} \\
\midrule
\multirow{4}{*}{NTU}
& RGB baseline & 0.391 & 0.243 \\
& LoRA (supervised) & 0.401 & 0.246 \\
& Supervised adapter & 0.413 & 0.257 \\
& \adapter\ (pose-guided) & \textbf{0.438} & \textbf{0.278} \\
\bottomrule
\end{tabular}
\end{table}

\section{Supporting Downstream Study: Social Attribute Prediction}
\label{sec:downstream}

A natural follow-up question is whether the improved behavior predictions carry information useful for downstream social interpretation. Recent work has shown that multimodal features extracted from video, audio, and transcripts can support automated assessments of high-level interaction qualities, including encouragement, warmth, and broader dimensions of teaching effectiveness~\cite{hou2024automated,futterer2026validating}. We treat this as supporting evidence rather than a primary benchmark, given the small dataset (204 dyadic ratings, 22 recordings) and context-dependent nature of the attributes.
 
BBSI was recorded from group interactions in which each participant rated other participants on rapport, perceived dominance, leadership, competence, and liking after the session~\cite{bernieri1996dyad, sanchez2011nonverbal}. These ratings provide a direct test: If \adapter\ produces more accurate behavior predictions, and if those behaviors genuinely encode social information, then using \adapter's predictions as features should yield better social attribute estimates than using the baseline's predictions.
 
\subsection{Setup}
\label{sec:downstream_setup}
For each subject in each recording, we aggregate clip-level behavior predictions into a 14-dimensional frequency vector (proportion of clips predicting each behavior). For each dyadic rating (subject $i$ rates subject $j$), we use the vector of ratings for the rated subject as features and the rating as the target. We fit Ridge regression ($\alpha = 1.0$) under leave-one-recording-out CV (22 folds) and report Spearman $\rho$ between predicted and actual scores. We use rank correlation because the ratings are ordinal. We assess significance with a paired permutation test (10,000 iterations) on $\Delta\rho = \rho_{\text{PRIMA}} - \rho_{\text{baseline}}$. We compare two feature sources: RGB baseline and \adapter\ predictions, both from Qwen3~VL. No social labels are used during adapter training. 

\begin{figure}[t]
\centering
\includegraphics[width=\columnwidth]{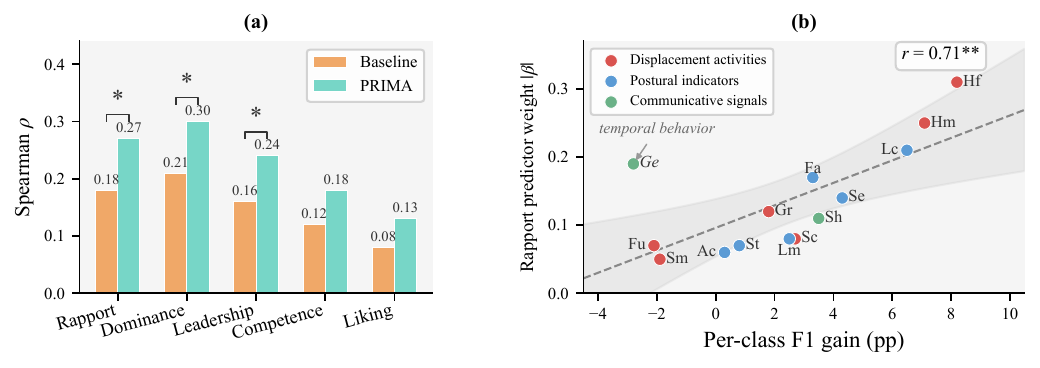}
\Description{Left: grouped bar chart comparing Spearman correlations for five social attributes between RGB baseline and PRIMA. Right: scatter plot of per-class F1 gain against absolute regression coefficient in the rapport model, showing positive correlation with confidence band.}
\caption{Downstream social attribute prediction on BBSI (Qwen3~VL, Ridge, leave-one-recording-out CV). (a)~\adapter\ improves correlation with human ratings on all five attributes; $*$: $p<0.05$, paired permutation test. (b)~Per-class F1 gain correlates with each behavior's weight in the rapport regression ($r=0.71$, $p<0.01$). Hf: Hand-face, Hm: Hand-mouth, Lc: Legs crossed, Fa: Fold arms, Se: Settle, Sh: Shrug, Ge: Gesture, Gr: Groom, Sc: Scratch, Lm: Leg movement, St: Stretching, Ac: Adjusting clothing, Fu: Fumble, Sm: Smearing hands.}
\label{fig:downstream}
\end{figure}

\subsection{Results}
\label{sec:downstream_results}
\Cref{fig:downstream}a shows that \adapter's behavior features produce higher correlations than the baseline on all five attributes, significantly on dominance ($\rho$: $0.21 \to 0.30$, $p = 0.02$), rapport ($0.18 \to 0.27$, $p = 0.03$), and leadership ($0.16 \to 0.24$, $p = 0.04$). Competence and liking improve ($+0.06$, $+0.05$) but do not reach significance, consistent with prior work showing these constructs depend more on facial expression than body configuration~\cite{tickle1990nature}. The absolute correlations are modest, as expected from zero-shot binary predictions. Applying the same Ridge predictor to ground truth behavior annotations yields $\rho \approx 0.36$ on rapport, against 0.27 for \adapter\ and 0.18 for the baseline, so better recognition alone can close only about half of the remaining gap. A permutation control and L1 normalization both confirm that the gains are not artifacts.

\subsection{Behavior-Level Attribution}
\label{sec:downstream_attribution}
The overall correlations do not indicate which specific recognition improvements are responsible. To answer this, we extract the standardized Ridge regression coefficients for each of the 14 behavior predictors in the rapport model and plot them against \adapter's per class F1 gain (\Cref{fig:downstream}b). The scatter reveals a positive association ($r = 0.71$, $p < 0.01$): The behaviors \adapter\ improves most are also the strongest predictors of rapport. The three behaviors with the largest F1 gains, Hand-face (+8.2\,pp), Hand-mouth (+7.1\,pp), and Legs crossed (+6.5\,pp), are classified as displacement activities or closed postures in the Ethological Coding System for Interviews~\cite{troisi1999ethological} that underlies BBSI's annotation. Displacement activities are established negative correlates of rapport~\cite{troisi1999ethological, knapp1972nonverbal, goodwin2011automated}; their regression coefficients are strongly negative ($\beta = -0.31, -0.25, -0.21$), so more accurate detection translates directly into better rapport prediction. Gesture ($-$2.8\,pp) carries a positive rapport coefficient ($\beta = +0.19$) but depends on temporal trajectory, so the spatial prior does not lead to improved predictions.

\section{Conclusion}
\label{sec:conclusion}

We proposed \adapter, a lightweight adapter that uses human pose as a training-time teaching signal to inject a body-structure prior into frozen vision-language models, with no pose input at inference. Trained on a single micro-action dataset, the prior improves zero-shot recognition on its source domain, transfers to two unseen target taxonomies, and can be partially reused across a second VLM backbone with the shared Core frozen. The transfer is selective: spatial body-part relations benefit most, fine hand detail and temporal trajectory least. 

In a supporting downstream study on BBSI, \adapter's predictions correlate more strongly with human ratings of rapport, dominance, and leadership, suggesting that better zero-shot behavior predictions can serve as a useful intermediate signal for social interpretation.

\noindent\textbf{Limitations.}
The prior is bounded by what the skeleton represents: major body joints are tracked, but hand and finger detail is absent, and each frame is transformed independently without cross-frame dynamics. We evaluate reuse across two additional VLM backbones at 8B and 4B scale within the same architecture family; transfer across architectural families and outdoor or unconstrained settings remains untested. The absolute zero-shot scores remain lower than those of supervised methods trained on target data, and studying \adapter\ as an initialization for few-shot adaptation is a natural next step.

\begin{acks}
This work was funded by the Deutsche Forschungsgemeinschaft (DFG, German Research Foundation) under project number 563097140 as part of the Priority Programme SPP 2481 UGAZE (Understanding Gaze).
\end{acks}

\clearpage
\bibliographystyle{ACM-Reference-Format}
\balance
\bibliography{sample-base}

\end{document}